\documentclass[12pt]{article}
\usepackage{times}
\usepackage{graphicx}
\usepackage[square,sort,comma,numbers]{natbib}
\title{Alternating Decision trees for early diagnosis of dengue fever}
 \author{M.~Naresh Kumar, \\
  \multicolumn{1}{p{.7\textwidth}}{\centering\emph{National Remote Sensing Centre (ISRO),
  India, nareshkumar\_m@nrsc.gov.in}}}

\begin{document}
\maketitle

\begin{abstract}
Dengue fever is a flu-like illness spread by the bite of an infected mosquito which is fast emerging as a major health problem. Timely and cost effective diagnosis using clinical and laboratory features would reduce the mortality rates besides providing better grounds for clinical management and disease surveillance. We wish to develop a robust and effective decision tree based approach for predicting dengue disease. Our analysis is based on the clinical characteristics and laboratory measurements of the diseased individuals. We have developed and trained an alternating decision tree with boosting and compared its performance with C4.5 algorithm for dengue disease diagnosis. Of the $65$ patient records a diagnosis establishes that $53$ individuals have been confirmed to have dengue fever. An alternating decision tree based algorithm was able to differentiate the dengue fever using the clinical and laboratory data with number of correctly classified instances as 89\%,  F-measure of $0.86$ and receiver operator characteristics (ROC) of $0.826$ as compared to C4.5 having correctly classified instances as $78\%$,h F-measure of 0.738 and ROC of 0.617 respectively. Alternating decision tree based approach with boosting has been able to predict dengue fever with a higher degree of accuracy than C4.5 based decision tree using simple clinical and laboratory features. Further analysis on larger data sets is required to improve the sensitivity and specificity of the alternating decision trees.
\end{abstract}


\section{Introduction}
Dengue fever is a mosquito-borne infectious disease and is re-emerging worldwide and causing larger and more frequent epidemics, especially in cities in the tropics and has become a major international public health concern. Dengue is found in tropical and sub-tropical regions around the world, predominantly in urban and semi-urban areas. The disease is caused by four distinct, but closely related viruses which are transmitted to humans through the bites of infective female Aedes mosquitoes \citep{1}. Recovery from infection by one provides life long immunity against that virus but confers only partial and transient protection against subsequent infection by the other three viruses. There is good evidence that sequential infection increases the risk of developing a more acute form of the disease known as dengue hemorrhagic fever (DHF) and dengue shock syndrome (DSS) which can be fatal. The mortality rate ranges from 6 to 30 percent, most commonly in children. The main pathophysiology of DHF and DSS is the development of plasma leakage from the capillaries, resulting in haemoconcentration, asciteps, and pleural effusion that may lead to shock following defervescence of fever \citep{2}. There is no vaccine yet for DF/DHF and management of the cases is largely supportive \citep{3}.

\begin{figure}[htbp]
	\centering
		\includegraphics[width=1.00\textwidth]{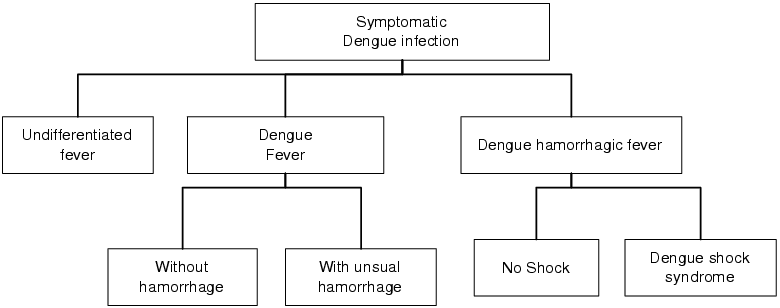}
		\caption{WHO Classification of Dengue fever into dengue fever, dengue hemorrhagic fever and dengue shock syndrome}
	\label{fig:figure1}
\end{figure}

Dengue illness is often confused with other viral febrile states confounding both clinical management (\citep{4,5,6}) and disease surveillance for viral transmission prevention \citep{7}. These difficulties especially strike during the early phase illness, wherein specific clinical symptoms and signs accompany the febrile illness \citep{4}. More definitive symptoms such as retro-orbital pain and rashes do not appear until the later stages of illness. Therefore a definitive early diagnosis requires laboratory tests such as Enzyme-linked immunosorbent assays (ELISA) and RT-PCR, which is based on nucleic and acid hybridization  (\citep{12,10,2}). Further the places where dengue is endemic lack the necessary infrastructure to carry out these tests \citep{7}, thus a scheme for reliable clinical diagnosis based on the data that can be obtained routinely, would be useful for early recognition of dengue fever. The current World Health Organization (WHO) scheme for classifying dengue infection (Fig.~\ref{fig:figure1}) makes use of symptoms which are not often present in the early stages of the infection, and thus it is not useful for early diagnosis.

The univarite and multivariate statistical techniques can provide a list of symptoms and signs based on clinical and laboratory features that can be associated with dengue (\citep{8,9}), but does not throw any light on diagnosis of the disease. Evidence-based triage strategies that identify individuals likely to be in the early stages of dengue illness are needed to be developed to help directing patient stratification for clinical investigations, management, and virological surveillance. To address this concern by utilizing alternating decision tree procedure \citep{14} which can generate more accurate, smaller and easier classification rules to interpret, when compared with decision trees such as C4.5 \citep{13} for diagnosing dengue fever.

\section{Material and Methods}
The clinical records consist of the following clinical features such as fever duration (FD), vomiting/nausea, body pains, rashes, pulse, headache, restlessness, abdominal pain) and following laboratory features such as hemoglobin (HB), white blood cells (WBC), platelet (PLT), packed cell volume (PCV), immunoglobin M (IgM) and immunoglobin G (IgG). Missing values in the continuous variables are replaced with their mean value (Table. \ref{tab:Table2}) before subjecting to further analysis.  We refer our readers to \citep{25} for a more detailed treatment on missing values in databases. As more than 40\% of the clinical records have missing values for the attributes IgG and IgM, they are excluded from our analysis.

\begin{table}[htbp]
	\centering
		\begin{tabular}{l|l}
		\hline
			FD	&1 value -- replaced with 7.4844 \\
Pulse	&1 value -- replaced with 89.6719\\
HB	&7 values -- replaced with 12.0224\\
WBC	&8 values -- replaced with 9.4684\\
PLT	&7 values -- replaced with 194.6207\\
PCV	&24 values -- replaced with 43.0488\\
\hline

		\end{tabular}
		\caption{Missing values for continuous attributes in the dataset is replaced with mean value} \label{tab:Table2}
\end{table}

To ascertain the clinical and laboratory features important in diagnosing the disease, continuous variables are subject to multinomial logistic regression implemented in TANAGRA data mining software \citep{22}. Using the Wald statistic (Table. \ref{tab:Table3}) variables significant for diagnosis (HB, WBC, Pulse), is selected.

\begin{table}[htbp]
	\centering
		\begin{tabular}{l|l|l}
		\hline
			Attribute &$\chi^2$ wald &	p-value \\
		\hline
Pulse&	0.692	&0.4056 \\
HB&	7.239	&0.0071 \\
WBC&	1.642&	0.2000 \\
PLT	&0.451	&0.5018 \\
PCV	&0.504	&0.4777 \\
\hline

		\end{tabular}
	\caption{Multinominal Logistic Regression to Select significant continuous variables}
	\label{tab:Table3}
\end{table}

The discrete/categorical variables are subjected to Chi Squared test procedure implemented in Tanagra (Table. \ref{tab:Table4}) to select attributes showing significant impact on the analysis.

\begin{table}[htbp]
	\centering
		\begin{tabular}{l|l|l|l}
		\hline
		Attribute	& $\chi^2$&	p-value	&95\% C.I. \\
		\hline
Vomiting/Nauseia&	0.94&	0.3320&	-0.0440 ; 0.0743\\
BodyPains	&0.17	&0.6837	&-0.0228 ; 0.0281\\
Rashes	&1.52	&0.4675	&-0.0419 ; 0.0908\\
Bleedingsite&	0.01	&0.9272&	-0.0057 ; 0.0060\\
Headache	&6.73	&0.0095	&-0.0497 ; 0.2661\\
Restlessness&	0.24	&0.6228	&-0.0281 ; 0.0358\\
Abdominal Pain&	1.05	&0.3063	&-0.0486 ; 0.0823\\
\hline

		\end{tabular}
	\caption{Chi squared statistics for categorical variables}
	\label{tab:Table4}
\end{table}

To evaluate the performance of the decision trees we have used popular classification measures such as sensitivity, specificity, precision, recall, F-measure and receiver operator characteristics (ROC) in our analysis. The definitions of the above measures are discussed for the benefit of the readers. Sensitivity measures the proportion of actual positives which are correctly identified as positives (TP). Specificity measures the proportion of negatives which are correctly identified as negatives. A theoretical, optimal prediction can achieve 100\% sensitivity (i.e. predict all people from the sick group as sick) and 100\% specificity (i.e. not predict anyone from the healthy group as sick). The ROC is a plot between sensitivity and 1-specificity is related in a direct and natural way to cost/benefit analysis (\citep{23,24}) of diagnostic decision making and provides the most comprehensive description of diagnostic accuracy available to date, since it estimates and reports all of the combinations of sensitivity and specificity that a diagnostic test is able to provide. F-measure is a weighted harmonic mean of precision and recall, and measures the effectiveness of retrieval with respect to a user who attaches $\beta$ times as much importance to recall as precision.

\section{Alternating Decision Trees for Diagnosis of Dengue Fever} 
Alternating decision trees (ADTrees) are machine learning methods combining boosting and decision trees algorithms to generate classification rules \citep{14}. Traditional boosting decision tree algorithms such as CART \citep{15} and C4.5 \citep{16} have been successful in generating classifiers but at the cost of creating complicated decision-tree structures. Such structures often represent convoluted decision rules that are hard to interpret \citep{14}. In contrast, ADTrees generate simpler decision-tree structures and easy-to interpret classification rules. ADTrees, are natural extensions of both voted-stumps and decision trees, consist of alternating layers of prediction and decision nodes \citep{14}. We refer our readers to \citep{26} for a detailed explanation of how ADTree is generated. The structure of an ADTree represents decision paths; when a path reaches a decision node, it continues with the specific offspring node that corresponds to the decision outcome as in the standard decision tree. On the other hand, when a path reaches a prediction node, the path continues with all of the offspring nodes. Thus the classification rule that it represents is basically a weighted majority vote over base prediction rules.

Boosting is a general and effective method of combining moderately successful rules to produce a very accurate prediction. Each weak prediction rule in the AdaBoost algorithm (\citep{17,18}) is associated with a prediction node. At each boosting iteration step, t, a decision node, together with its two prediction nodes, is introduced. For full ADTrees, the decision node may be attached to any previous prediction node, leaf nodes or otherwise, including the root prediction node. Each prediction node is associated with a weight, $\alpha$ which represents its contribution to the final prediction score, F(x), for every path that reaches it. Hence the contribution of each decision node may be understood in isolation, and summing the individual contributions gives rise to the final prediction and classification.

\begin{figure}[htbp]
	\centering
		\includegraphics[width=1.00\textwidth]{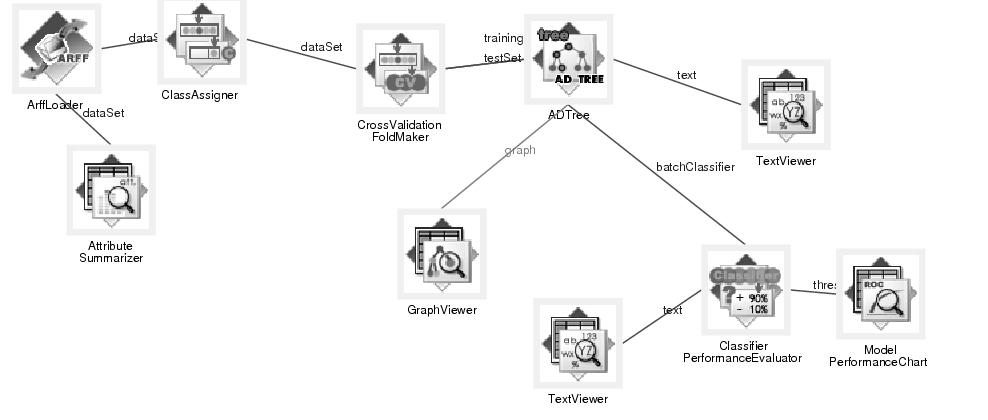}
	\caption{Knowledge flow layout in Weka for training ADTree}
	\label{fig:figure2}
\end{figure}

The clinical and laboratory observations of the dengue disease are used in generating alternating decision trees and J48 (implementation of C4.5) using Weka 3.6.1 (\citep{21} open source tool for data mining). The knowledge flow layout for ADTree training in Weka is given in Fig.~\ref{fig:figure2}. A k-fold cross validation approach is adopted for testing the predictions as it is considered to be a powerful methodology to overcome data over fitting \citep{20}. The fold value is set to k=10 which is the nominal value used for cross fold validation \citep{19}. The ADTree outputs are measured using Graph Viewer component and the classification accuracy can be seen in the text viewer. The ADTree classifier is evaluated using the performance evaluator and the graphs can be visualized using model performance chart component. 

\section{Results}
\begin{figure}[htbp]
	\centering
		\includegraphics[width=1.00\textwidth]{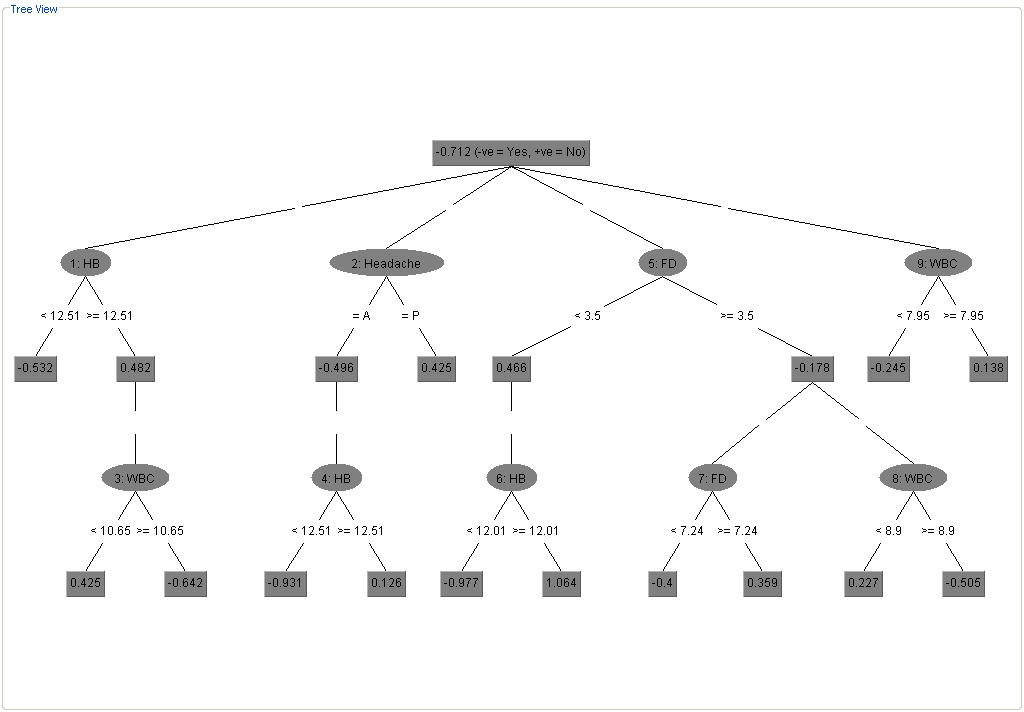}
	\caption{ADTree generated after converting pulse attribute into binary variable -VE = YES, +VE = NO}
	\label{fig:figure3}
\end{figure}

\begin{figure}[htbp]
	\centering
		\includegraphics[width=1.00\textwidth]{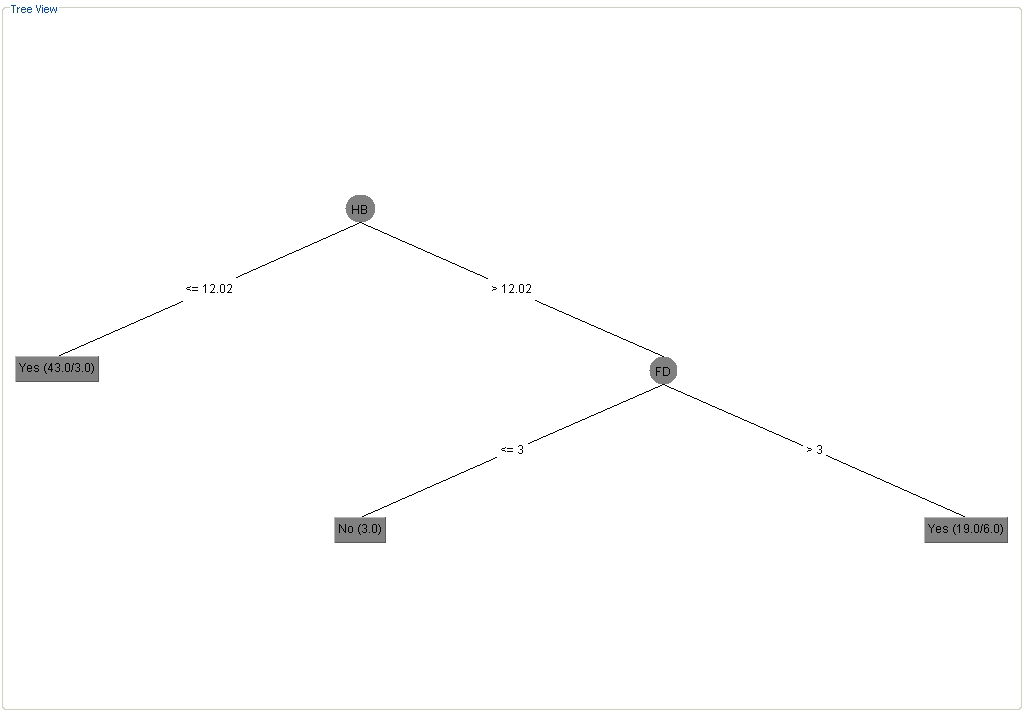}
	\caption{Decision tree generated using J48 in Weka with discretized values of Pulse variable}
	\label{fig:figure4}
\end{figure}

The ADTree and J48 decision trees are shown in respectively in Fig.~\ref{fig:figure3} and Fig.~\ref{fig:figure4}. The ADTree with attributes FD, pulse, HB, and WBC as continuous variables, headache as a discrete variable with cross fold validation (k=10) was able to correctly classify 84\% of the cases, whereas the J48 was able to classify only 78\% of cases correctly.

The number of correctly classified instances using ADTree improved to 89\% after adopting the method of discretizing \citep{9}  the pulse ( pulse <100  Low (L), otherwise High (H) ) attributes.
The confusion matrix is given in Table \ref{tab:Table5} suggests a classification accuracy of 100\% for all positively diagnosed cases (true positive).  The negative cases diagnosed as positive (false positive) are more in J48 than in ADTree approach indicating a better performance of ADTrees. The false positives identified in ADTrees are further analyzed and it was found that some instances have values replaced earlier for missing values.

\begin{table}[htbp]
	\centering
		\begin{tabular}{l|l|l|l}
		\hline
	&&&			Predicted \\
	\hline
Algorithm	&Actual&	Yes&	No \\
\hline
J48	&Yes&	50	&3 \\
	& &	(TP)&(FN)\\
 &&94\%	&5\% \\
			\hline
&	No	&11&	1 \\
		&&(FP)&	(TN)\\
		&&91.60\%&	8.30\% \\
	\hline
ADTree&	Yes&	53&	0 \\
		&&(TP)&	(FN)\\
		&&100\%&	0\%\\
		\hline
	&No&	7	& 5\\
		&&(FP)	&(TN)\\
		&&58.30\%&	41.60\% \\
\hline	
		\end{tabular}
	\caption{Confusion matrix for J48 AND ADTree}
	\label{tab:Table5}
\end{table}

The weighted average values of TP rate, FP rate, F-measure and ROC over positive and negative outcomes of the diagnosis for the ADTree and J48 decision trees are tabulated in Table \ref{tab:Table6}.

\begin{table}[htbp]
	\centering
		\begin{tabular}{l|l|l|l|l}
		\hline
			Algorithm	& Sensitivity	& 1-Specificity	& F-Measure &	ROC \\
			\hline
ADTree&	0.892	&0.476&	0.873	&0.826 \\
J48	&0.785	&0.758	&0.738	&0.617 \\
\hline
		\end{tabular}
	\caption{Performance measures forJ48 AND ADTree classifiers}
	\label{tab:Table6}
\end{table}

\begin{figure}[htbp]
	\centering
		\includegraphics[width=1.00\textwidth]{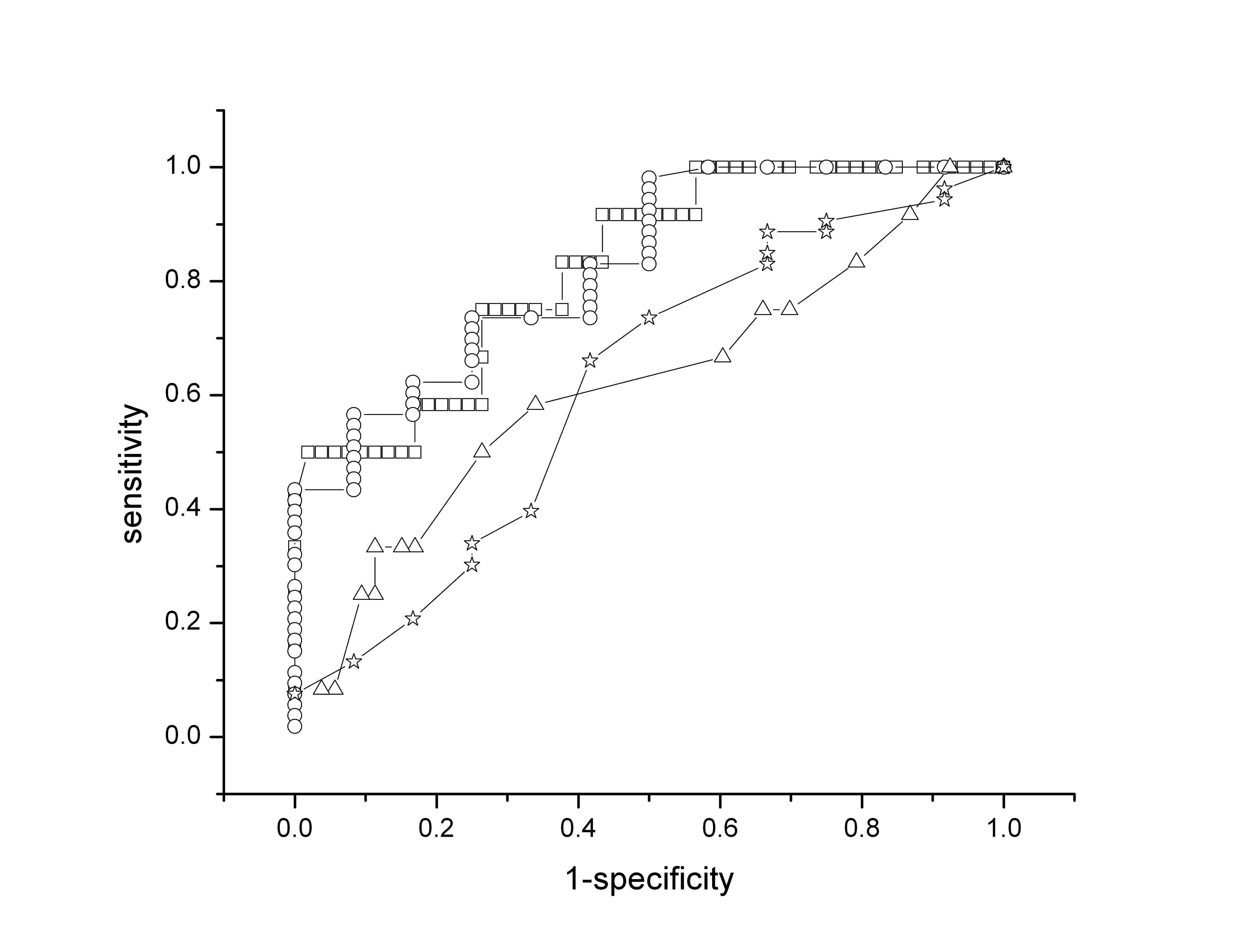}
	\caption{Sensitivity and Specificity curves for ADTree}
	\label{fig:figure5}
\end{figure}

A sensitivity of 89\% for a specificity of 52\% is achieved using ADTree (Fig.~\ref{fig:figure5}) with only 65 datasets, without considering the IgM and IgG antibody attributes, has performed better in comparison to logistic regression approach (\citep{9}, sensitivity 90\% and specificity 58\% for 412 patients) and C4.5 decision tree approach (\citep{13} ROC AUC 0.82 for 1012 cases).

\section{Discussion}
Using an alternating decision tree algorithm with boosting for analysis of all clinical and hematological data, we obtained diagnostic rules that discriminates dengue from non-dengue illness with an accuracy of 84\% and improved classification of 89\% when the attribute pulse is converted into a categorical variable. The specificity in ADTree can be further improved by providing a sufficient number of examples of non-dengue cases. This study shows a proof-of-concept that alternating decision trees with boosting using simple clinical and laboratory parameters can predict the diagnosis of dengue disease, a finding that could prove useful in disease management and surveillance.
\bibliographystyle{plainnat}

\end{document}